# Nested Dictionary Learning
# for Hierarchical Organization of Imagery and Text

**Lingbo Li**
Duke University
Durham, NC 27708
ll83@duke.edu

**XianXing Zhang**
Duke University
Durham, NC 27708
xz60@duke.edu

**Mingyuan Zhou**
Duke University
Durham, NC 27708
mz31@duke.edu

**Lawrence Carin**
Duke University
Durham, NC 27708
lcarin@duke.edu

## Abstract

A tree-based dictionary learning model is developed for joint analysis of imagery and associated text. The dictionary learning may be applied directly to the imagery from patches, or to general feature vectors extracted from patches or superpixels (using any existing method for image feature extraction). Each image is associated with a path through the tree (from root to a leaf), and each of the multiple patches in a given image is associated with one node in that path. Nodes near the tree root are shared between multiple paths, representing image characteristics that are common among different types of images. Moving toward the leaves, nodes become specialized, representing details in image classes. If available, words (text) are also jointly modeled, with a path-dependent probability over words. The tree structure is inferred via a nested Dirichlet process, and a retrospective stick-breaking sampler is used to infer the tree depth and width.

## 1 Introduction

Statistical topic models, such as latent Dirichlet allocation (LDA) (Blei et al., 2003b), were originally developed for text analysis, but they have been recently transitioned successfully for the analysis of imagery. In most such topic models, each image is associated with a distribution over topics, and each topic is characterized by a distribution over observed features in the image. In this setting researchers typically represent an image as a bag of visual words (Fei-Fei & Perona, 2005; Li & Fei-Fei, 2007). These methods have been applied to perform unsupervised clustering, classification and annotation of images, using image features as well as auxiliary data such as image annotations (Barnard et al., 2003; Blei & Jordan, 2003; Blei & MaAuliffe, 2007; Wang et al., 2009; Li et al., 2009).

In such work feature extraction is performed as a pre-processing step, and local image descriptors, *e.g.*, scale-invariant feature transform (SIFT) (Lowe, 1999), and other types of features (Arbelaez & Cohen, 2008), are commonly used to extract features from local patches (Fei-Fei & Perona, 2005; Li & Fei-Fei, 2007; Bart et al., 2008; Sivic et al., 2008; Wang et al., 2009), segments (Li et al., 2009; Yang et al., 2010), or superpixels (Du et al., 2009). Different images are related to one another by their corresponding distributions over topics.

There are several limitations with most of this previous work. First, vector quantization (VQ) is typically applied to the image features (Fei-Fei & Perona, 2005), and the codes play the role of "words" in traditional topic modeling. There is a loss of information in this quantization step, and one must tune the number of codes (the proper codebook may change with different types of images, and as new imagery are observed). Secondly, feature design is typically performed separately from the subsequent image topic modeling. Finally, most of the image-based topic modeling is performed at a single scale or level (Wang et al., 2009; Li et al., 2009; Du et al., 2009), thereby not accounting for the hierarchical characteristics of most natural imagery.

In recent work, there have been papers that have addressed particular aspects of the above limitations, but none that has addressed all. For example, in Li et al. (2011) the authors employed a dictionary-learning framework, eliminating the need to perform VQ. This dictionary learning could be applied to traditional features pre-computed from the image, or it could be applied directly to patches of raw imagery, thereby ameliorating the requirement of separating the feature-design and topic-modeling steps. However, Li et al. (2011) did not consider the hierarchical charac-

ter of imagery. Recently Li et al. (2010) employed a nested Chinese restaurant process (nCRP) to infer a hierarchical tree representation for a corpus of images and (if available) accompanying text; however, in that work the VQ step was still employed, and therefore a precomputation of features was as well. Further, in Li et al. (2010), while the tree width was inferred, the depth was set. Finally, the nCRP construction in Li et al. (2010) has the disadvantage of only updating parent-child-transition parameters from one node of the tree at a time, in a sampler, yielding poor mixing relative to a stick-breaking Dirichlet process (DP) implementation (Ishwaran & James, 2001). Related but distinct dictionary learning with the nCRP was considered in Zhang et al. (2011).

Motivated by these recent contributions, and the limitations of most existing topic models of imagery and text, this paper makes the following contributions:

- A nested DP (nDP) model is developed to learn a hierarchical tree structure for a corpus of imagery and text, with a stick-breaking construction employed; we infer both the tree depth and width, using a retrospective stick-breaking construction (Papaspiliopoulos & Roberts, 2008).

- A beta-Bernoulli dictionary learning framework (Zhou et al., 2011b) is adapted to such a hierarchical model, removing the VQ step, and allowing one to perform topic modeling directly on image patches, thereby integrating feature design and topic modeling. However, if desired, the dictionary learning may also be applied to features pre-computed from the image, using *any* existing method for feature design, and again removing the limitations of VQ.

## 2 Modeling Image Patches

We wish to build a hierarchical model to arrange $M$ images and their associated annotations (when available); the vocabulary of such annotations is assumed to be of dimension $N_v$. The vector $\boldsymbol{x}_{mi}$ represents the pixels or features associated with the $i$th patch in image $m$, and $\boldsymbol{y}_m = (y_{m1}, \ldots, y_{mN_v})^T$ represents a vector of word counts associated with that image, when available ($y_{mn}$ represents the number of times word $n \in \{1, \ldots, N_v\}$ is present in the annotation).

The $m$th image is divided into $N_m$ patches (or superpixels (Li et al., 2010)), and the data for the $i$th patch is denoted $\boldsymbol{x}_{mi} \in \mathbb{R}^P$ with $i = 1, \ldots, N_m$. The vector $\boldsymbol{x}_{mi}$ may represent raw pixel values, or a feature vector extracted from the pixels (using any available method of image feature extraction, *e.g.*, SIFT (Lowe, 1999)).

Each $\boldsymbol{x}_{mi}$ is represented as a sparse linear combination of learned dictionary atoms. Further, each patch is assumed associated with a "topic"; the probability of which dictionary atoms are employed for a given patch is dictated by the topic it is associated with.

Specifically, each patch is represented as $\boldsymbol{x}_{mi} = \mathbf{D}(\boldsymbol{z}_{mi} \odot \boldsymbol{s}_{mi}) + \boldsymbol{e}_{mi}$, where $\odot$ represents the element-wise/Hadamard product, $\mathbf{D} = [\boldsymbol{d}_1, \cdots, \boldsymbol{d}_K] \in \mathbb{R}^{P \times K}$, $K$ is the truncation level on the possible number of dictionary atoms, $\boldsymbol{z}_{mi} = [z_{mi1}, \cdots, z_{miK}]^T$, $\boldsymbol{s}_{mi} = [s_{mi1}, \cdots, s_{miK}]^T$, $z_{mik} \in \{0, 1\}$ indicates whether the $k$th atom is *active* within patch $i$ in image $m$, $s_{mik} \in \mathbb{R}^+$, and $\boldsymbol{e}_{mi}$ is the residual. Note that $\boldsymbol{z}_{mi}$ represents the specific sparseness pattern of dictionary usage for $\boldsymbol{x}_{mi}$. The hierarchical form of the model is

$$\begin{aligned}
\boldsymbol{x}_{mi} &\sim \mathcal{N}(\mathbf{D}(\boldsymbol{z}_{mi} \odot \boldsymbol{s}_{mi}), \gamma_e^{-1} \mathbf{I}_P) \\
\boldsymbol{d}_k &\sim \mathcal{N}(0, \frac{1}{P} \mathbf{I}_P) \\
\boldsymbol{s}_{mi} &\sim \mathcal{N}_+(0, \gamma_s^{-1} \mathbf{I}_K) \\
\boldsymbol{z}_{mi} &\sim \prod_{k=1}^{K} \text{Bernoulli}(\pi_{h_{mi}k})
\end{aligned} \quad (1)$$

where gamma priors are placed on both $\gamma_e$ and $\gamma_s$. Positive weights $\boldsymbol{s}_{mi}$ (truncated normal, $\mathcal{N}_+(\cdot)$) are imposed, which we have found to yield improved results.

The indicator variable $h_{mi}$ defines the topic associated with $\boldsymbol{x}_{mi}$. The $K$-dimensional vector $\boldsymbol{\pi}_h$ defines the probability that each of the $K$ columns of $\mathbf{D}$ is employed to represent topic $h$, where the $k$th component of $\boldsymbol{\pi}_h$ is $\pi_{hk}$. These probability vectors are drawn

$$\boldsymbol{\pi}_h \sim G_0, \quad G_0 = \prod_{k=1}^{K} \text{Beta}(a_0/K, b_0(K-1)/K) \quad (2)$$

where $\pi_{hk}$ represents the probability of using $\boldsymbol{d}_k$ for object type $h$, and the introduction of $G_0$ is for discussions below. This representation for $\boldsymbol{\pi}_h$ corresponds to an approximation to the beta-Bernoulli process (Thibaux & Jordan, 2007; Paisley & Carin, 2009; Zhou et al., 2011a,b), which also yields an approximation to the Indian buffet process (IBP) (Griffiths & Ghahramani, 2005; Teh et al., 2007).

## 3 Tree Structure via nDP

The nested Dirichlet process (nDP) tree construction developed below is an alternative means of constituting the same type of tree manifested by the nested Chinese restaurant process (Blei et al., 2003a; Li et al., 2010). We emphasize the nDP construction because of the stick-breaking implementation we employ, which allows block updates, and therefore often manifests better mixing than the nCRP-type implementation (Ishwaran & James, 2001). Related work was considered in Wang & Blei (2009), but VB inference was employed and the tree size was therefore not inferred (a fixed truncation was imposed). The retrospective sampler developed below allows inference of both the tree depth and width (Papaspiliopoulos & Roberts, 2008).

Consider a draw from a DP, $G \sim \text{DP}(\gamma, G_0)$, where $\gamma > 0$ is an "innovation" parameter, with $G_0$ defined in (2). Then the DP draw (Ishwaran & James, 2001) may be expressed as $G = \sum_{n=1}^{\infty} \lambda_n \delta_{\phi_n}$, where $\lambda_n = \nu_n \prod_{l<n}(1-\nu_l)$, $\nu_l \sim \text{Beta}(1, \gamma)$, and $\phi_n \sim G_0$; each $\phi_n$ corresponds to a topic, as in (2). Letting $\boldsymbol{\lambda} = (\lambda_1, \lambda_2, \ldots)^T$, we denote the draw of $\boldsymbol{\lambda}$ as $\boldsymbol{\lambda} \sim \text{Stick}(\gamma)$.

### 3.1 Tree width

Using notation from Adams et al. (2010), let $\boldsymbol{\epsilon}$ represent a path through the tree, characterized by a sequence of parent-child nodes, and let $|\boldsymbol{\epsilon}|$ be the length of this path (total number of layers traversed). In addition to representing a path through the tree, $\boldsymbol{\epsilon}$ identifies a node at layer $|\boldsymbol{\epsilon}|$, i.e., the node at the end of path $\boldsymbol{\epsilon}$. For node $\boldsymbol{\epsilon}$, let $\boldsymbol{\epsilon}\epsilon_i$, $i = 1, 2, \ldots$, denote the *children* of $\boldsymbol{\epsilon}$, at level $|\boldsymbol{\epsilon}|+1$. To constitute a distribution over the children nodes, we draw $G_{\boldsymbol{\epsilon}} \sim \text{DP}(\gamma, G_0)$, yielding $G_{\boldsymbol{\epsilon}} = \sum_{i=1}^{\infty} \lambda_{\boldsymbol{\epsilon}\epsilon_i} \delta_{\phi_{\boldsymbol{\epsilon}\epsilon_i}}$, where $\lambda_{\boldsymbol{\epsilon}\epsilon_i} = \nu_{\boldsymbol{\epsilon}\epsilon_i} \prod_{j=1}^{i-1}(1-\nu_{\boldsymbol{\epsilon}\epsilon_j})$, $\nu_{\boldsymbol{\epsilon}\epsilon_j} \sim \text{Beta}(1, \gamma)$, and $\phi_{\boldsymbol{\epsilon}\epsilon_i} \sim G_0$, with $G_0$ defined in (2); $\boldsymbol{\lambda}_{\boldsymbol{\epsilon}} = (\lambda_{\boldsymbol{\epsilon}\epsilon_1}, \lambda_{\boldsymbol{\epsilon}\epsilon_2}, \ldots)^T$ is denoted as drawn $\boldsymbol{\lambda}_{\boldsymbol{\epsilon}} \sim \text{Stick}(\gamma)$. The probability measure $G_{\boldsymbol{\epsilon}}$ constitutes in principle an infinite set of children nodes, with $\lambda_{\boldsymbol{\epsilon}\epsilon_i}$ defining the probability of transiting from node $\boldsymbol{\epsilon}$ to child $\epsilon_i$; $\phi_{\boldsymbol{\epsilon}\epsilon_i}$ constitutes the topic-dependent probability of dictionary usage at that child node.

The process continues *in principle* to an infinite number of levels, with each child node spawning an infinite set of subsequent children nodes, manifesting a tree of infinite depth and width. However, note that a draw $\boldsymbol{\lambda}_{\boldsymbol{\epsilon}}$ will typically only have a relatively small number of components with appreciable amplitude. This means that while $G_{\boldsymbol{\epsilon}}$ constitutes in principle an infinite number of children nodes, only a small fraction will be visited with appreciable probability.

Let $\boldsymbol{c}_m = (c_m^1, c_m^2, \ldots)^T$ represent the path associated with image $m$, where $c_m^l$ corresponds to the node selected at level $l$. For conciseness we write $\boldsymbol{c}_m \sim \text{nCRP}(\gamma)$ (Blei et al., 2003a), emphasizing that the underlying transition probabilities $\boldsymbol{\lambda}_{\boldsymbol{\epsilon}}$, controlling the probabilistic path through the tree, are a function of parameter $\gamma$.

### 3.2 Tree depth

We also draw an associated probability vector $\boldsymbol{\theta}_m \sim \text{Stick}(\alpha)$. Patch $\boldsymbol{x}_{mi}$ is associated with level $l_{mi}$ in path $\boldsymbol{c}_m$, where $l_{mi} \sim \sum_{l=1}^{\infty} \theta_{ml} \delta_l$. Since $\boldsymbol{\theta}_m$ typically only has a small number of components with appreciable amplitude, the tree depth is also constrained.

### 3.3 Modeling words

In Section 2 we developed topic (node) dependent probabilities of atom usage; we now extend this to words (annotations), when available. A distribution over words may be associated with each topic (tree node) $h$. For topic $h$ we may draw (Blei et al., 2003b)

$$\boldsymbol{\psi}_h \sim \text{Dir}(\frac{\eta}{N_v}, \ldots, \frac{\eta}{N_v}) \quad (3)$$

where $\boldsymbol{\psi}_h$ is the distribution over words for topic $h$.

Recall that each image/annotation is associated with a path $\boldsymbol{c}_m$ through a tree, and $\boldsymbol{\theta}_m$ controls the probability of employing each node (topic) on that path. Let $\theta_{mh}$ represent the probability that node $h$ is utilized, with $h \in \boldsymbol{c}_m$. Then the "collapsed" probability of word usage on this path, marginalizing out the probability of node selection, may be expressed as

$$\boldsymbol{\psi}_{\boldsymbol{c}_m} = \sum_{h \in \boldsymbol{c}_m} \theta_{mh} \boldsymbol{\psi}_h \quad (4)$$

A probability over words $\boldsymbol{\psi}_{\boldsymbol{c}_m}$ is therefore associated with each path $\boldsymbol{c}_m$. One may argue that the $\boldsymbol{\theta}_m$ used to control node usage for image patches should be different from that used to represent words; this is irrelevant in the final model, as a path-dependent $\boldsymbol{\psi}_{\boldsymbol{c}_m}$ is drawn directly from a Dirichlet distribution (discussed below), and therefore (4) is only illustrative/motivating.

### 3.4 Retrospective sampling

The above discussion indicated that while the width and depth of the tree is infinite in principle, a finite tree is manifested given finite data, which motivates adaptive inference of the tree size. In a retrospective implementation of a stick-breaking process, we constitute a *truncated* stick-breaking process, denoted $\boldsymbol{w} \sim \text{Stick}_L(\gamma)$, with $w_n = V_n \prod_{l<n}(1-V_l)$, $V_n \sim \text{Beta}(1, \gamma)$ for $n < L$, and $V_L = 1$; here there is an $L$-stick truncation, yielding $\boldsymbol{w} = (w_1, \ldots, w_L)^T$, with $w_L$ representing the probability of selecting a stick *other* than sticks 1 through $L-1$.

In a retrospective sampler (Papaspiliopoulos & Roberts, 2008), each of the aforementioned sticks is truncated as above. When drawing children nodes and levels, if the last stick (the $L$th above) is selected, this implies that a new child/level must be added, since the first $L-1$ sticks are not enough to capture how the data are clustered. If stick $L$ is selected, then a new node/level is constituted (a new child is added), by drawing a new $V_L \sim \text{Beta}(1, \gamma)$, and then $V_{L+1} = 1$, thereby now constituting an $(L+1)$-dimensional stick representation; the associated node-dependent statistics are constituted as discussed in Section 2 (drawing new probabilities over dictionary elements). The model therefore infers "in retrospect" that the $L$-level truncation was too small, and expands adaptively. The model also has the ability to shrink the number of sticks used at any component of the model, if less than the associated truncated level is needed to define the number of children/levels are actually utilized.

### 3.5 Generative Process

The generative process for the model is summarized as follows:

1. Draw dictionary $\mathbf{D} \sim \prod_{k=1}^{K} \mathcal{N}(0, \frac{1}{P}\mathbf{I}_P)$

2. Draw $\gamma$, $\alpha$, $\gamma_e$ and $\gamma_s$ from respective gamma distributions

3. For each image $m \in \{1, 2, ..., M\}$
   (a) Draw $\mathbf{c}_m \sim \text{nCRP}(\gamma)$
   (b) For each *newly utilized* node $\boldsymbol{\epsilon}$ in the tree, draw dictionary usage probabilities $\boldsymbol{\pi}_{\boldsymbol{\epsilon}} \sim \prod_{k=1}^{K} \text{Beta}(a_0/K, b_0(K-1)/K)$
   (c) Draw $\boldsymbol{\theta}_m \sim \text{Stick}(\alpha)$
   (d) For the $i$th patch or feature vector
      i. Draw level index $l_{mi} \sim \sum_{l=1}^{\infty} \theta_{ml}\delta_l$, which along with $\mathbf{c}_{mi}$ defines node $h_{mi}$
      ii. Draw $\mathbf{z}_{mi} \sim \prod_{k=1}^{K} \text{Bernoulli}(\pi_{h_{mi}k})$, and $\mathbf{s}_{mi} \sim \mathcal{N}_+(0, \gamma_s^{-1}\mathbf{I}_K)$
      iii. Draw $\mathbf{x}_{mi} \sim \mathcal{N}(\mathbf{D}(\mathbf{z}_{mi} \odot \mathbf{s}_{mi}), \gamma_e^{-1}\mathbf{I}_P)$

4. For each unique tree path $p$, draw $\boldsymbol{\psi}_p \sim \text{Dir}(\frac{\beta}{N_V}, \ldots, \frac{\beta}{N_V})$

5. If annotations are available for image $m$, $\mathbf{y}_m \sim \text{Mult}(|\mathbf{y}_m|, \boldsymbol{\psi}_{\mathbf{c}_m})$, where $|\mathbf{y}_m|$ is the total number of words in $\mathbf{y}_m$

In Step 3(b), new nodes (topics) are added "in retrospect", as discussed in the previous subsection (nodes may also be pruned with this sampler). After completing Step 3, the tree size is constituted, which allows Step 4, imposition of a distribution over words for each path.

---

**Algorithm 1** Retrospective Sampling for $l_{mi}$
**Input:** $L_m$, $\mathbf{z}$, $\mathbf{l}$, $a_0$, $b_0$
**Output:** $l_{mi}$, $L_m$
**for** $m = 1$ **to** $M$ and $i = 1$ **to** $N_m$ **do**
  Sample $\mu_{m|\boldsymbol{\epsilon}|}$, $\boldsymbol{\pi}_{\boldsymbol{\epsilon}}$ from the conditional posterior for $|\boldsymbol{\epsilon}| \leq L_m$, and from the prior for $|\boldsymbol{\epsilon}| > L_m$;
  $\theta_{m|\boldsymbol{\epsilon}|} = \mu_{m|\boldsymbol{\epsilon}|} \prod_{s=1}^{|\boldsymbol{\epsilon}|-1}(1-\mu_{ms})$
  Sample $U_{mi} \sim \text{Uniform}[0, 1]$
  **if** $\sum_{s=1}^{j-1} q(l_{mi} = s) < U_{mi} \leq \sum_{s=1}^{j} q(l_{mi} = s)$ **then**
    Set $l_{mi} = j$ with probability $\kappa_{mi}(j)$, otherwise, leave $l_{mi}$ unchanged
  **else**
    $L_m = L_m + 1$, set $l_{mi} = L_m$ with probability $\kappa_{mi}(L_m)$, otherwise, leave $l_{mi}$ unchanged
  **end if**
**end for**

---

## 4 Model Inference

A contribution of this paper concerns use of retrospective sampling to infer the tree width and depth. To save space for an extensive set of experimental results, we here only discuss updates associated with inferring the tree depth. A complete set of update equations are provided in Supplementary Material, where one may also find a summary of all notation.

To sample $l_{mi}$ from the conditional posterior, we first need to specify the likelihood that $\{\boldsymbol{\epsilon} \in \mathbf{c}_m\}$:

$$p(\mathbf{z}_{mi}|\boldsymbol{\pi}_{\boldsymbol{\epsilon}}, \mathbf{c}_m) = \prod_{k=1}^{K} \pi_{\boldsymbol{\epsilon}k}^{z_{mik}}(1-\pi_{\boldsymbol{\epsilon}k})^{1-z_{mik}}$$

and the prior distribution, which is specified by a stick-breaking draw $\boldsymbol{\theta}_m$ for each image $m$. Although $l_{mi}$ can be sampled from a closed form posterior for a fixed $L_m$, here to learn $L_m$ adaptively we instead use an Metropolis-Hastings step, where the proposal distribution is defined as

$$q(l_{mi} = j) \propto \begin{cases} \theta_{mj} p(\mathbf{z}_{mi}|\boldsymbol{\pi}_j, \mathbf{c}_m), & j \leq L_m \\ \theta_{mj} \mathcal{M}_{mi}(L_m), & j > L_m \end{cases}$$

where $\mathcal{M}_{mi}(L_m) = \max_{1 \leq |\boldsymbol{\epsilon}| \leq L_m} \{p(\mathbf{z}_{mi}|\boldsymbol{\pi}_{\boldsymbol{\epsilon}}, \mathbf{c}_m)\}$. Note that the sampled value of $l_{mi}$ is allowed to be larger than the truncation level $L_m$, consequently $L_m$ and the depth of the tree is learned adaptively. The acceptance probability $\kappa_{mi}(j)$ for $l_{mi} = j$ is

$$\begin{cases} 1, & j \leq L_m \ \& \ L'_m = L_m \\ \min\{1, \frac{\tilde{c}_{mi}(L_m)\mathcal{M}_{mi}(L'_m)}{\tilde{c}_{mi}(L'_m)p(\mathbf{z}_{mi}|\boldsymbol{\pi}_{h_{mi}}, \mathbf{c}_m)}\}, & j \leq L_m \ \& \ L'_m < L_m \\ \min\{1, \frac{\tilde{c}_{mi}(L_m)p(\mathbf{z}_{mi}|\boldsymbol{\pi}_j, \mathbf{c}_m)}{\tilde{c}_{mi}(L'_m)\mathcal{M}_{mi}(L_m)}\}, & j > L_m \end{cases}$$

where the normalizing constant is defined as $\tilde{c}_{mi}(L_m) = \sum_{|\boldsymbol{\epsilon}|=1}^{L_m} \theta_{m|\boldsymbol{\epsilon}|} p(\mathbf{z}_{mi}|\boldsymbol{\pi}_{\boldsymbol{\epsilon}}, \mathbf{c}_m) + \mathcal{M}_{mi}(L_m)(1-$

$\sum_{|\boldsymbol{\epsilon}|=1}^{L_m} \theta_{m|\boldsymbol{\epsilon}|})$. The retrospective sampling procedure for $l_{mi}$ is summarized in Algorithm 1.

## 5 Experiments

We test the proposed model with five datasets: $(i)$ a simulated, illustrative example that examines the ability to learn the tree structure; $(ii)$ a subset of the MNIST digits data, $(iii)$ face data (Tenenbaum et al., 2000); $(iv)$ the Microsoft (MSRC) image database; and $(v)$ the LabelMe data. In the case of $(iv)$ and $(v)$, the images are supplemented by annotations. For $(ii)$-$(v)$, we process patches from each image. For the MNIST and face data, we randomly select 50 partially overlapping patches in each image, with $15 \times 15$ and $40 \times 40$ patch sizes, respectively (placement of patches was not tuned, selected uniformly at random, with partial overlap). For the MSRC and LabelMe data, we collect all $32 \times 32 \times 3$ non-overlapping patches from the *color* images (we also consider overlapping patches in this case, but it was found unnecessary). Recall that $\boldsymbol{x}_{mi} \in \mathbb{R}^P$, with $P$ the number of pixels in each patch ($P = 225$ for MNIST, $P = 1600$ for the face data, and $P = 3072$ for MSRC and LabelMe data).

We have examined different methods for initializing the dictionary, including random draws from the prior and various fixed redundant bases, such as the over-complete DCT. Alternatively, we may use existing dictionary-learning methods (independent of the topic model); for this purpose, we use the covariate-dependent hierarchical beta process (with the covariates linked to the relative locations between patches) to learn an initial set of dictionary atoms (Zhou et al., 2011b). Additionally, in examples $(ii)$-$(v)$, we initialize the tree with 4 levels. In this initialization, four nodes are present beneath the root node, and each subsequent node has two children, down four levels; nested K-means clustering is used to initialize the data among the clusters (nodes) at the respective levels.

In all experiments, the hyperparameters were set $a_0 = b_0 = 1$, $c_0 = d_0 = e_0 = f_0 = 10^{-6}$, $\alpha = 1$ and $\gamma = 1$, and the truncation level (upper bound) on the number of dictionary elements was $K = 400$; many related settings of these parameters yield similar results, and no tuning was performed.

### 5.1 Inferring the tree: simulated data

We first illustrate that the proposed model is able to infer both the depth and width of the tree, using synthesized data for which the tree that generates the data is known; the data are like (but distinct from) that considered in Figure 2 of Blei et al. (2003a). In this simple example we wish to isolate the component of the model that infers the tree, so there is no dictionary learning. The data consists of a 25-element alphabet, arranged as $5 \times 5$ blocks on a grid; each topic is characterized by the probability of using each of the 25 elements (there is a probability $\pi_{hk}$ for using element $k \in \{1, \ldots, 25\}$, for each topic/node $h$). For each topic $h$ the "truth" (middle in Figure 1) has probabilities 0 (blue), 0.1 (cyan blue), and 0.5 (red). The generative process for image $m$ corresponds to first drawing a path $\boldsymbol{c}_m$ through the tree, and then 1000 times a node on this patch is drawn from $\boldsymbol{\theta}_m$, and finally each of the 25 alphabet members are drawn Bernoulli, with the associated topic-dependent probabilities (this is the proposed model, without dictionary learning). The final data ("image") consists of a count of the number of times each of the 25 elements was used, across the 1000 draws (example data at left in Figure 1). A total of 100 $5 \times 5$ "images" were drawn in this manner to constitute the data. The right part of Figure 1 corresponds to the *recovered* tree, based upon the maximum-likelihood collection sample. Of course, the order of the branches and children is arbitrary; the inferred tree in Figure 1 (right) was arranged *a posteriori* to align with the "truth" (middle), to clarify presentation.

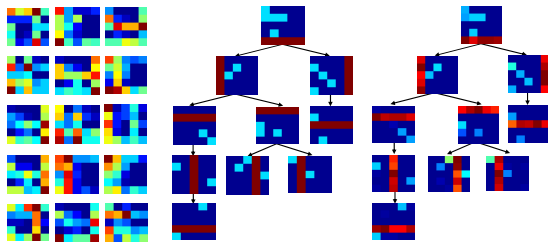

Figure 1: Example with synthesized images. Left: Example images. Middle: The ground truth for the underlying model. Right: The inferred model from the maximum-likelihood collection sample.

In this example the tree was initialized with 3 paths, each with three 3 layers (note in truth there are four paths, with variable number of layers). We also initialized the tree with 4 and 5 paths, under the same experimental setting, and similar recovery is achieved. If we initialized with less than 3 paths or more than 5, the recovered tree was still reasonable (close), but not as good. However, the inference of the topic-dependent usage probabilities $\boldsymbol{\pi}_h$ was very robust to numerous different settings. These results are based on 2000 samples, 1000 discarded as burn-in.

### 5.2 Model fit

For the MNIST handwritten digit database, we randomly choose 100 images per digit (digits 0 through

9), and therefore $M = 1000$ images are considered in total; the images are of size $28 \times 28$. The face dataset (Tenenbaum et al., 2000) contains $M = 698$ faces, each of size $64 \times 64$. Concerning the inferred trees, for the MNIST data, the maximum-likelihood collection sample had 168 paths and each path was typically 5 layers deep; for the face data 80 paths were inferred, and each was typically 5 layers. To quantitatively compare the ability of the hierarchical dictionary construction to fit the data, we consider reconstruction error for the data, comparing with the single-layer ("flat") model in Li et al. (2011); results are summarized in Table 1, corresponding to $\|\boldsymbol{x}_{mi} - \mathbf{D}(\boldsymbol{z}_{mi} \odot \boldsymbol{s}_{mi})\|_2^2$, averaged across all images $m$ and patches $i$. In addition to results for MNIST and faces data, we show results for the MSRC and LabelMe data sets (analyzed in this case *without* annotations).

We also performed experiments to investigate how initialization affects the performance. In Table 1, instead of initializing the dictionary via the hierarchical beta process (hBP), they are initialized at random. While there is a slight degradation in performance with random initialization, it is not marked, and the results are still better than those produced by Li et al. (2011). Similar improvements in hBP initialization were observed for the classification task discussed below; hBP helps, but random initialization is still good.

Table 1: Reconstruction error comparison (mean square error multiplied by $10^3$, and $\pm$ one standard deviation) on MNIST, Face, MSRC and LabelMe datasets. 'nDP+hBP' and 'nDP+random' correspond to the proposed model with the dictionary initialized by hBP and randomly, respectively, while the "flat" (single layer) model corresponds to Li et al. (2011).

|  | MNIST | Face | MSRC | LabelMe |
|---|---|---|---|---|
| flat model | 11.10 ± 0.32 | 8.18 ± 0.17 | 10.27 ± 0.54 | 12.16 ± 0.30 |
| nDP+hBP | 10.42 ± 0.21 | 7.64 ± 0.12 | 8.64 ± 0.36 | 10.21 ± 0.27 |
| nDP+random | 10.85 ± 0.35 | 7.91 ± 0.20 | 9.25 ± 0.41 | 11.53 ± 0.50 |

The proposed model fits the data better than the "flat" model in Li et al. (2011); the gains are more evident when considering real and sophisticated imagery (the MSRC and LabelMe data). It is important to note that the proposed model is effectively no more complicated than the model in Li et al. (2011). Specifically, in Li et al. (2011) and in the proposed model, each image is put in a cluster, where in Li et al. (2011) a single-layer Dirichlet process was used to perform clustering, where here paths through the tree define clusters. In Li et al. (2011) and here each cluster/path is characterized by a distribution over topics, and in both models each topic is characterized by probabilities of atom usage (and each model has a truncation level on the dictionary of the same size). The difference is that in Li et al. (2011) the probabilities of topic usage for each cluster are drawn independently, where here the tree structure, and shared nodes between different paths, manifest statistical dependencies between the probability of topic usage in different paths with shared nodes.

In these examples a total of 250 Gibbs samples were run, with 150 discarded as burn-in. The results of the model correspond to averaging across the collection samples. In all examples useful results were found with a relatively small number of Gibbs samples.

### 5.3 Organizing MSRC data

We use the same settings of images and annotations from the MSRC data[1] as considered in Du et al. (2009), to allow a direct comparison. We choose 320 images from 10 categories of images with manual annotations available. The categories are "tree", "building", "cow", "face", "car", "sheep", "flower", "sign", "book" and "chair". The numbers of images are 45 and 35 in the "cow" and "sheep" classes, respectively, and 30 in all the other classes. Each image has size $213 \times 320$ or $320 \times 213$. For annotations, we remove all words that occur less than 8 times (approximately 1% of words), and obtain 15 unique annotation-words, thus $N_v = 15$.

The full tree structure inferred is shown in Figure 2, with its maximum depth inferred to be 6 (maximum-likelihood collection sample depicted, from 100 collection samples). On the second level, it is clearly observed that images are clustered in several main subgenres, *e.g.*, one with images containing grass, one with flowers, and another with urban construction, including cars and buildings, etc. For each path, we depict up to the 8 most-probable images assigned to it (fewer when less than 8 were assigned).

To further demonstrate the form of the model, two pairs of example images are shown in Figure 3. The pairs of images were assigned to two distinct paths, that shared nodes near the root (meaning the model infers shared types of patches between the images, assigned to these shared nodes). For each node, three example patches that are assigned to it are selected, from each of the two images. For the pair of example images from the "sheep" and "cow" classes, patches of grass and legs are shared on the top nodes, while distinct patches manifesting color and texture are separately assigned to nodes at bottom levels. The two images from the "building" class show the diversity of this category. It is anticipated that the "building" category will be diverse, with common patches shared at nodes near the root, and specialized patches near the leaves (for different types of buildings/structures). These typical examples illustrate that ubiquitous patches are shared at nodes near the root, with nodes toward the leaves describing details associated with specialized classes of images. It is this hierarchical structure that is missed by the model in Li et al. (2011), and that also apparently manifests the better model fit, as summarized in Table 1.

---
[1] http://research.microsoft.com/en-us/projects/objectclassrecognition/

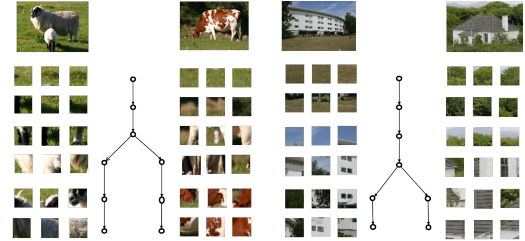

Figure 3: Two pairs of example images, and the paths they were assigned to. Between the images is shown the splitting paths. At top are the example images, and for each image we depict three example patches assigned to a respective node.

Another product of the model is a distribution over words for each path in the tree (not shown, for brevity). We illustrate this component of the model for the LabelMe data, considered next.

### 5.4 Organizing LabelMe data

The LabelMe data[2] contain 8 image classes: "coast", "forest", "highway", "inside city", "mountain", "open country", "street" and "tall building". We use the same settings of images and annotations as Wang et al. (2009): we randomly select 100 images for each class, thus the total number of images is 800. Each image is resized to be $256 \times 256$ pixels. For the annotations, we remove terms that occur less than 10 times, and obtain a vocabulary of 99 unique words, thus $N_v = 99$. There are 5 terms per annotation in the LabelMe data on average. Figure 6 visualizes 7 sub-trees of the inferred tree structure; there are 7 nodes inferred on the second level, and each node represents one sub-tree. Class "street" and class "insidecity" share the same root node, labeled 5 in Figure 6.

Based on the learned posterior word distribution $\boldsymbol{\psi}_p$ for the $p$th image class, we can further infer which words are most probable for each path. Figure 4 shows the $\boldsymbol{\psi}_p$ for 8 example paths (maximum-likelihood sample, from 100 collection samples), with the five largest-probability words displayed; the capital letters associated with each histogram in Figure 4 have associated paths through the tree as indicated in Figure 6. A good connection is manifested between the words and paths (examine the images and words associated with each path).

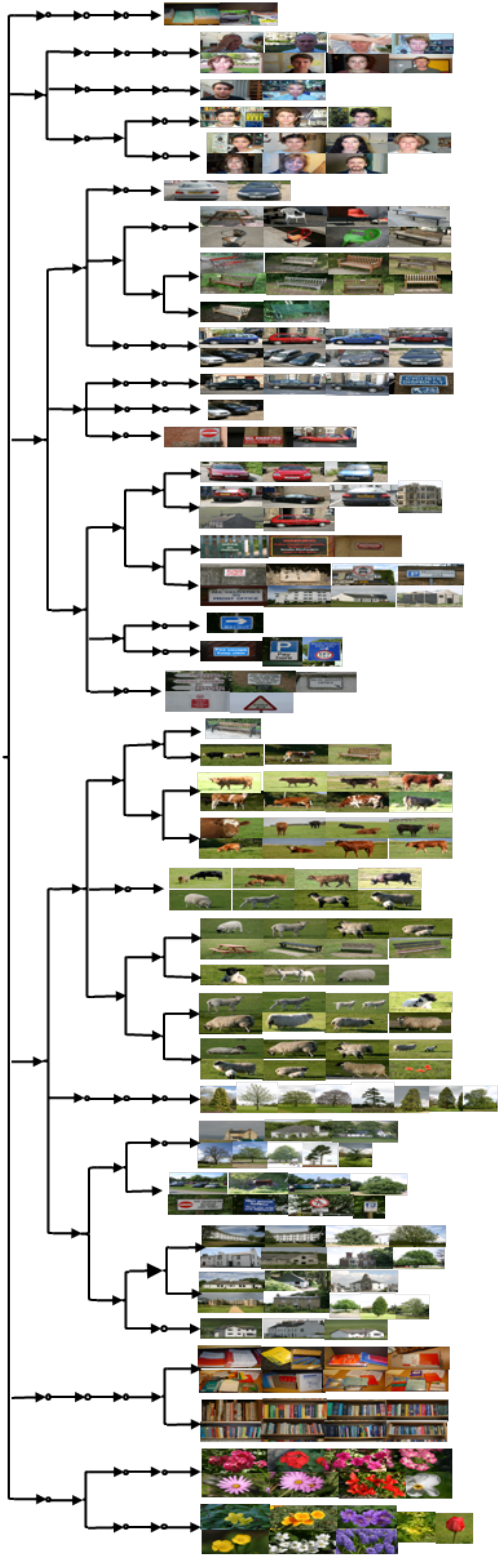

Figure 2: The full tree structure inferred from MSRC data. For each path, up to 8 images assigned to it are shown.

---
[2]http://www.cs.princeton.edu/~chongw/

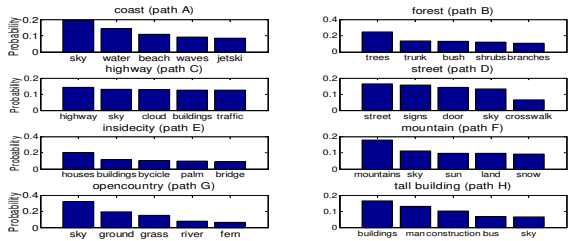

Figure 4: Inferred distributions over words for LabelMe data, as a function of inferred image category. The letters correspond to paths in Figure 6.

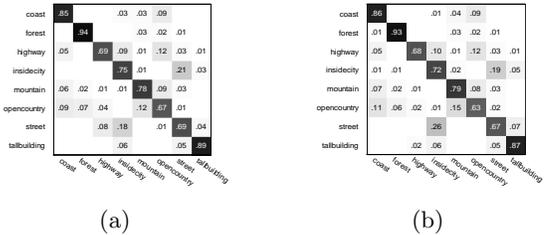

Figure 5: For the 800 testing images from LabelMe data, (a): Confusion Matrix on original patches with the average accuracy of 78.3%. (b): Confusion Matrix on SIFT features with the average accuracy of 76.9%.

We evaluate the proposed model on the image-classification task, similar to as considered in Li et al. (2010). A set of 800 randomly selected images are held out as testing images from the 8 classes, each class with 100 testing images. Each image is represented by the estimated distribution over all the nodes in the entire hierarchy. Only nodes that are associated to the image have nonzero values in the distribution. We calculate the $\chi^2$-distances between the node distribution of the testing images and those of the training images. The KNN algorithm ($K$ is set to be 50) is then applied to obtain the class label.

Figure 5(a) shows the confusion matrix of classification, with an average classification accuracy of 78.3%, compared with 76% in Li et al. (2011). In all of the above examples the dictionary learning was applied directly to the observed pixel values within a given patch, with no *a priori* feature extraction. Alternatively, the patch-dependent data $x_{mi}$ may correspond to features extracted using *any* image feature extraction algorithm. To illustrate this, we now let $x_{mi}$ correspond to SIFT features (Lowe, 1999) on the same patches; in this experiment the dictionary learning replaces the VQ step in models like Wang et al. (2009). In Figure 5(b) we show the confusion matrix of the model based on SIFT features, with an average accuracy of 76.9%, slightly better than the results reported in Wang et al. (2009) (but here there is no need to tune the number of VQ codes). This also demonstrates that performing dictionary learning directly on the patches, rather than via a state-of-the-art feature extraction method, yields highly competitive results.

We now compare the proposed hierarchical model with the hierarchical model in Li et al. (2010), in which offline SIFT feature extraction and VQ are employed. Based on related work in Wang et al. (2009), we used a codebook of size 240, and achieved an average classification accuracy of 77.4%, compared with 79.6% reported above for our algorithm. Note, however, that we found the model in Li et al. (2010) to be very sensitive to the codebook size, with serious degradation in performance manifested with 150 or 400 codes, for example. To further test the proposed model, we considered the same classification experiment on MSRC data, which is characterized by 10 classes. Five images per class were randomly chosen as testing data, and the remaining images are treated as training data to learn a hierarchical structure. An average accuracy of 64% is obtained with the proposed model, compared with 60% using that in Li et al. (2010), where the codebook size is set to be 200. These experiments indicate that the proposed model typically does better than that in Li et al. (2010) for the classification task, even when we optimize the latter with respect to the number of codes.

The experiments above have been performed in 64-bit Matlab on a machine with 2.27 GHz CPU and 4 Gbyte RAM. One MCMC sample of the proposed model takes approximately 4, 2, 8 and 10 minutes respectively for the MNIST, Face, MSRC and LabelMe experiments (in which we simultaneously analyzed respectively 1000, 698, 320, and 800 total images). Note that while these model *learning* times are relatively expensive, model testing (after the tree and dictionary are learned) is very fast, this employed for the aforementioned classification task. To scale the model up to larger numbers of training images, we may perform variational Bayesian inference rather than sampling, and employ online-learning methods (Hoffman et al., 2010; Carvalho et al., 2010).

## 6 Conclusions

The nested Dirichlet process has been integrated with dictionary learning to constitute a new hierarchical topic model for imagery. The dictionary learning may be employed on the original image pixels, or on features from any image feature extractor. If words are available, they may be utilized as well, with word-dependent usage probabilities inferred for each path through the tree. The model infers both the tree depth and width. Encouraging qualitative and quantitative results have been demonstrated for analysis of many of the traditional datasets in this field, with comparisons provided to other related published methods.

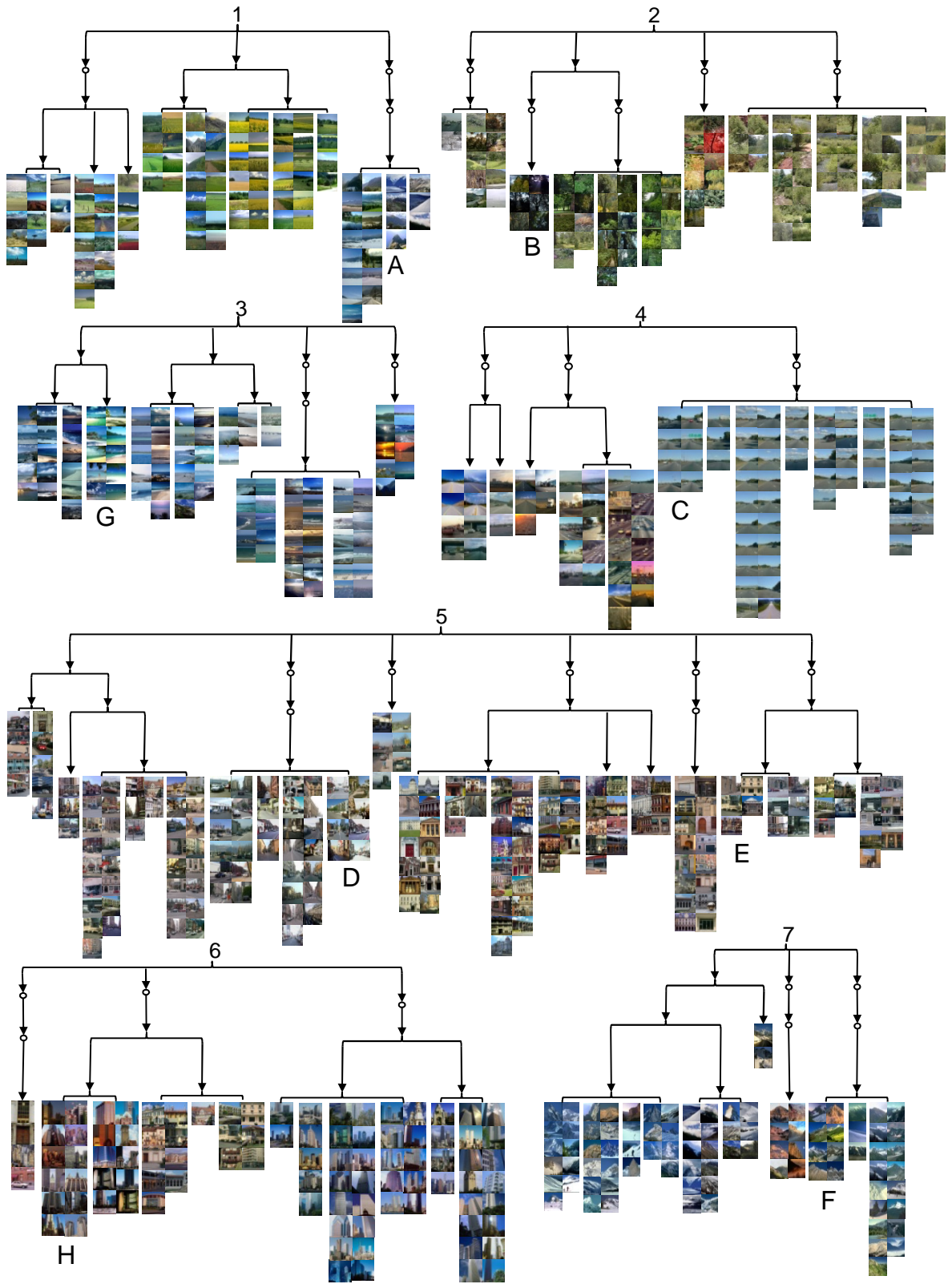

Figure 6: The structure of 7 sub-trees inferred from LabelMe data. The root of each sub-tree is one child node of the root for the entire tree structure. For each path, all images assigned to it are listed with no order importance. The letters refer to paths for which distributions over words are depicted in Figure 4.

## Acknowledgement

The work reported here was supported by ARO, DOE, NGA, ONR and DARPA (under the MSEE Program).